%% file: elsarticle-template-arxiv.tex
\PassOptionsToPackage{dvipsnames}{xcolor}
\documentclass[review]{elsarticle}
\makeatletter
\def\ps@pprintTitle{%
 \let\@oddhead\@empty
 \let\@evenhead\@empty
 \def\@oddfoot{}%
 \let\@evenfoot\@oddfoot}
\makeatother
\usepackage{tikz}
\usepackage{comment}
\usepackage{amsmath,amssymb} 
\usepackage{color}
\usepackage{epsfig}
\usepackage[inline]{enumitem}
\usepackage[caption=false]{subfig}
\usepackage{soul}
\usepackage[utf8]{inputenc} 
\usepackage[T1]{fontenc}    
\usepackage{booktabs}       
\usepackage{amsfonts}       
\usepackage{nicefrac}       
\usepackage{microtype}      
\usepackage{algorithmic}
\usepackage{algorithm}
\usepackage{numprint}
\usepackage{siunitx}
\usepackage{etoolbox}
\newcommand{\ubold}{\fontseries{b}\selectfont}
\robustify\ubold
\usepackage{bbold}
\usepackage{wrapfig}
\usepackage{bm}
\usepackage[caption=false]{subfig}
\usepackage{pifont}

\usepackage{lineno,hyperref}
\usepackage{array,multirow,graphicx}

\usepackage[dvipsnames]{xcolor}
\makeatletter
\newcommand\footnoteref[1]{\protected@xdef\@thefnmark{\ref{#1}}\@footnotemark}
\makeatother

\modulolinenumbers[5]

\journal{Journal of Pattern Recognition}

\begin{document}

\pagenumbering{gobble}

\let\WriteBookmarks\relax
\def\floatpagepagefraction{1}
\def\textpagefraction{.001}
\pagenumbering{arabic}

\input{math_def}
\newcommand{\cV}{\mathcal{V}}
\newcommand{\cE}{\mathcal{E}}
\newcommand{\cB}{\mathcal{B}}
\newcommand{\cG}{\mathcal{G}}
\newcommand{\etal}{\textit{et al.}}
\newcommand{\xmark}{\ding{55}}%
\newcommand{\cmark}{\ding{51}}%

\begin{frontmatter}

\title{ActivityCLIP: Enhancing Group Activity Recognition by Mining Complementary Information from Text to Supplement Image Modality}

\author[1]{Guoliang Xu}
\author[2]{Jianqin Yin\corref{cor1}}
\ead{jqyin@bupt.edu.cn}
\cortext[cor1]{Corresponding author}
\author[1]{Feng Zhou}
\author[1]{Yonghao Dang}

\address[1]{School of Artificial Intelligence, Beijing University of Posts and Telecommunications, Beijing, China}
\address[2]{School of Intelligent Engineering and Automation, Beijing University of Posts and Telecommunications, Beijing, China}


\begin{abstract}
Previous methods usually only extract the image modality's information to recognize group activity. However, mining image information is approaching saturation, making it difficult to extract richer information. Therefore, extracting complementary information from other modalities to supplement image information has become increasingly important. In fact, action labels provide clear text information to express the action's semantics, which existing methods often overlook. Thus, we propose ActivityCLIP, a plug-and-play method for mining the text information contained in the action labels to supplement the image information for enhancing group activity recognition. ActivityCLIP consists of text and image branches, where the text branch is plugged into the image branch (The off-the-shelf image-based method). The text branch includes Image2Text and relation modeling modules. Specifically, we propose the knowledge transfer module, Image2Text, which adapts image information into text information extracted by CLIP via knowledge distillation. Further, to keep our method convenient, we add fewer trainable parameters based on the relation module of the image branch to model interaction relation in the text branch. To show our method's generality, we replicate three representative methods by ActivityCLIP, which adds only limited trainable parameters, achieving favorable performance improvements for each method. We also conduct extensive ablation studies and compare our method with state-of-the-art methods to demonstrate the effectiveness of ActivityCLIP.
\end{abstract}
\end{frontmatter}

\section{Introduction}
Group Activity Recognition (GAR) is an important task in video understanding, which has been applied in many fields, such as surveillance, sports video analysis, and human behavior understanding \cite{Deng2022Summarization,ehsanpour2020joint,kim2018discriminative}. 

 \begin{figure}[htbp]
  \centering
  \includegraphics[width=0.7\textwidth]{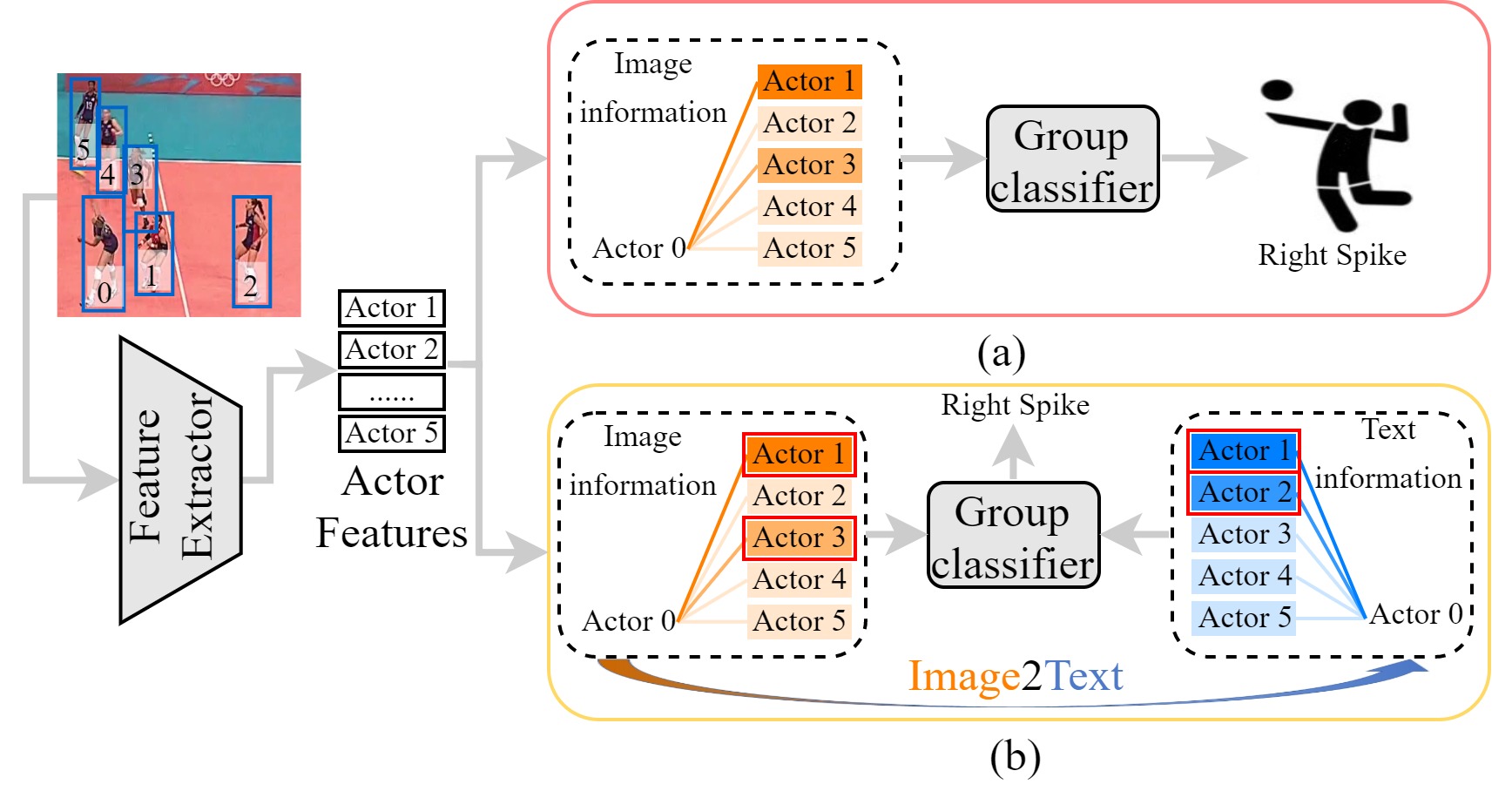}\\
  \caption{(a) Image-based group activity recognition; (b) Ours (image-text-based group activity recognition). Our method can mine complementary information (Deeper colors indicate stronger interaction with actor 0.) from the text to supplement image information to enhance group activity recognition.}\label{Figure 1}
 \end{figure}

Many methods have been proposed to only use image information directly for GAR \cite{wu2019learning,gavrilyuk2020actor,han2022dual,li2021groupformer,xie2023actor}, as shown in Fig. \ref{Figure 1}(a). Specifically, they used the backbone to extract the image features and further cropped the actor features from the image features by the RoIAlign \cite{he2017mask}. Finally, they captured the actor interaction relation based on actor features to recognize group activity. Clearly, the entire process exclusively extracts information from the image modality. However, mining image information is approaching saturation, making it difficult to extract more abundant information. Therefore, it is urgent to explore how to mine complementary information from other modalities to supplement image information for enhancing group activity recognition.

Due to the limited information contained in image modality, several methods have also been proposed to mine the information on skeleton modality \cite{perez2022skeleton,zhou2022composer,bian2022self,wang2024augmented}. Specifically, these methods usually use an off-the-shelf pose estimation method to extract the skeleton information from the image. Then, they model the actor interaction relation based on the skeleton information for recognizing group activity. However, these methods have two apparent disadvantages. First, the scene of group activity is usually crowded, and actors often cover each other. Thus, accurately estimating skeleton information in this scene is a challenging problem. Second, these methods are usually more complex because an additional stage for estimating skeleton information is needed. Unlike them, we propose to mine the text information to supplement the image information, as shown in Fig. \ref{Figure 1}(b).

In the process of mining text information, we need to consider two issues. First, we hope to excavate the text information without adding additional data and annotation. Second, we hope the text information is universal and can conveniently provide complementary information to various image-based methods.

To address the two issues, we propose a novel method, named ActivityCLIP, to recognize group activity. About the first issue, we find that action labels provide clear text information to express the action's semantics. Therefore, we propose the Image2Text module to mine the semantic information contained in the action labels. Specifically, in the training stage, we use CLIP's text encoder \cite{radford2021learning} to encode the action label's semantics to text information. Then, we use knowledge distillation to achieve information transfer between text information and the Image2Text's output features (The Image2Text module uses image information as input.). In this way, the Image2Text module can learn the information transfer capacity from image to text information. In the testing stage, we only use the Image2Text module rather than CLIP’s text encoder to obtain text information. About the second issue, ActivityCLIP consists of image and text branches, where we plug the text branch into the image branch. Specifically, we add fewer trainable parameters into the relation modeling module of the image branch and further train these parameters to model interaction relations in the text branch. Finally, we fuse the scores of both branches to recognize group activity.

In summary, the main contributions of our research are as follows.
\begin{itemize}
  \item We proposed a novel method, ActivityCLIP, that mines the text information contained in action labels to supplement image information for enhancing group activity recognition.
   
  \item Our method is a plug-and-play method that conveniently plugs the text branch into the image branch. In detail, we propose the Image2Text module that can transfer image information to text information. Further, we add fewer trainable parameters based on the image branch's relation modeling module to achieve the interaction relation modeling of the text branch.

  \item We replicate three representative methods to evaluate ActivityCLIP's generality. We find that ActivityCLIP can significantly improve the performance of these methods. Additionally, we conduct extensive ablation studies and compare our method with state-of-the-art methods, demonstrating its effectiveness.
\end{itemize}

\section{Related Works}
\subsection{Vision Language Model} 
Many VLMs (Vision Language Models) have demonstrated impressive performance in multimodality tasks \cite{radford2021learning,jia2021scaling,furst2022cloob,wang2021simvlm,desai2021virtex,li2023scaling,sun2023eva}. One of the influential works is CLIP, which belongs to the contrastive vision language model \cite{radford2021learning}. CLIP is trained using contrastive learning on a dataset of 400 million image-text pairs. CLIP contains an image encoder and a text encoder to extract the image and text features, respectively. It aims to maximize the similarity of the correct image-text pairs and minimize the similarity of the incorrect image-text pairs. A comprehensive review of VLM is beyond the scope of this paper, and the details can be referred to \cite{chen2023vlp,du2022survey}. Because of CLIP's impressive performance, we chose it to mine the text information. However, we need to seriously consider how to use CLIP in downstream tasks.

\subsection{CLIP Applied in Downstream Tasks} 
Finetuning the VLM with all parameters requires significant computation and data consumption. Thus, researchers hope to apply VLM to various downstream tasks \cite{rao2022denseclip,wang2023actionclip,zhang2023clamp,yuan2024open} by finetuning fewer or not finetuning parameters. 

CLIP has demonstrated the zero-shot ability in the image classification task. Then, Wang et al. extended CLIP to the video action classification task that they regarded as a video-text matching problem \cite{wang2023actionclip}. By adding the text and visual adapter to the text and visual encoder, respectively, their method not only has superior zero-shot/few-shot ability but also reaches significant performance. Zhang et al. applied CLIP to the pose estimation task and regarded it as a keypoint-text matching problem \cite{zhang2023clamp}. They encouraged the visual feature of specific keypoint to maximize the similarity to the text prompt describing the corresponding keypoint, while minimizing the similarity to the text prompt describing other keypoints. In this way, the text information can enhance the image information to locate the keypoints accurately. Rao et al. used CLIP to realize more complex dense prediction tasks, such as semantic and instance segmentation \cite{rao2022denseclip}. They regarded dense prediction tasks as a pixel-text matching problem to obtain the pixel-text score map. The score map contains the prior knowledge contained in CLIP and further guides the learning of dense prediction models.

We find an action-text matching problem naturally exists in our task. We maximize the similarity of the correct action-text pairs and minimize the similarity of incorrect action-text pairs. Through this process, we use CLIP’s text encoder to teach our model how to mine the aligned information with the text information, which enables our model to extract text and image information simultaneously.

\subsection{Group Activity Recognition} 

According to the information modality, we can divide the previous methods into image-based methods \cite{wu2019learning,gavrilyuk2020actor,han2022dual,li2021groupformer,xie2023actor,li2022learning} and skeleton-based methods \cite{perez2022skeleton,zhou2022composer,bian2022self}. The image-based methods aim to extract the image features, such as appearance, pose, motion features, etc, as the actor's representation. Wu et al. \cite{wu2019learning} proposed ARG (Actor Relation Graph), which used Inception-v3 \cite{szegedy2016rethinking} to extract the actor's image features and further built actor relation graphs to model the actor interaction relation for recognizing group activity. Like ARG, Han et al. \cite{han2022dual} proposed Dual-AI, which also used inception-v3 as the backbone. Further, they proposed the dual path with different spatial-temporal orders to mine the abundant image information. Different from ARG, Kirill et al. \cite{gavrilyuk2020actor} proposed AFormer (Actor transFormer), which used the more powerful backbone (HRNet \cite{sun2019deep} and I3D \cite{carreira2017quo}) to extract pose features and motion features simultaneously, attempting to get stronger image information. However, mining richer image information has become increasingly difficult. Therefore, it is crucial to introduce information from other modalities to assist the image modality.

Skeleton-based methods aim to estimate the human skeleton from the image and further use skeleton information to recognize group activity. Mauricio et al. \cite{perez2022skeleton} used OpenPose \cite{cao2017realtime} to extract skeleton information. Then, they modeled the intra-person, inter-person, and person-object relation only based on the skeleton information to recognize group activity. Zhou et al. \cite{zhou2022composer} also only used skeleton information to recognize group activity. They used HRNet to extract skeleton information and proposed to analyze the interaction relation from four scales. However, skeleton estimation is a challenging problem, especially in a crowded scene. Meanwhile, skeleton-based methods need to estimate the skeleton information in advance, which increases the complexity of GAR tasks. Therefore, we did not extract the skeleton information but mined the text information to supplement image information for recognizing group activity.

\section{Methodology}
\subsection{Preliminaries}
\textbf{CLIP.} We briefly introduce the content related to CLIP \cite{radford2021learning}. CLIP consists of an image encoder and a text encoder. It aims to align the image-text features in the latent space through contrastive learning. The input of CLIP is image-text pairs. The text encoder $ {E_t} $ is responsible for extracting text feature $ x $. The image encoder $ {E_i} $ is responsible for extracting image feature $ y $. As shown in Eq. (\ref{eq1}), CLIP hopes that the positive pair has a high probability and the negative pair has a low probability. \textit{Sim} is the cosine similarity, $\tau$ is the learnable temperature parameter, and \textit{K} is the number of training pairs. Finally, the symmetric CE (Cross-Entropy) loss is used to optimize the CLIP. \textit{G} is the ground truth.

\begin{equation} \label{eq1}
\begin{array}{l}
P_i^{x,y}(x) = \frac{{\exp (sim(x,{y_i})/\tau )}}{{\sum\nolimits_{j = 1}^K {\exp (sim(x,{y_j})/\tau )} }}\\
P_i^{y,x}(y) = \frac{{\exp (sim(y,{x_i})/\tau )}}{{\sum\nolimits_{j = 1}^K {\exp (sim(y,{x_j})/\tau )} }}
\end{array}
\end{equation}

\begin{equation} \label{eq2}
\begin{array}{l}
L = \frac{1}{2}[CE({P^{x,y}},{G^{x,y}}) + CE({P^{y,x}},{G^{y,x}})]
\end{array}
\end{equation}

Due to its impressive generality and multi-modality alignment ability, we use CLIP to teach our model how to mine text information. To avoid excessive resource consumption, we freeze the parameters of CLIP and adopt knowledge distillation to achieve knowledge transfer.

\begin{figure*}[htbp]
\centering
\includegraphics[width=1.0\textwidth]{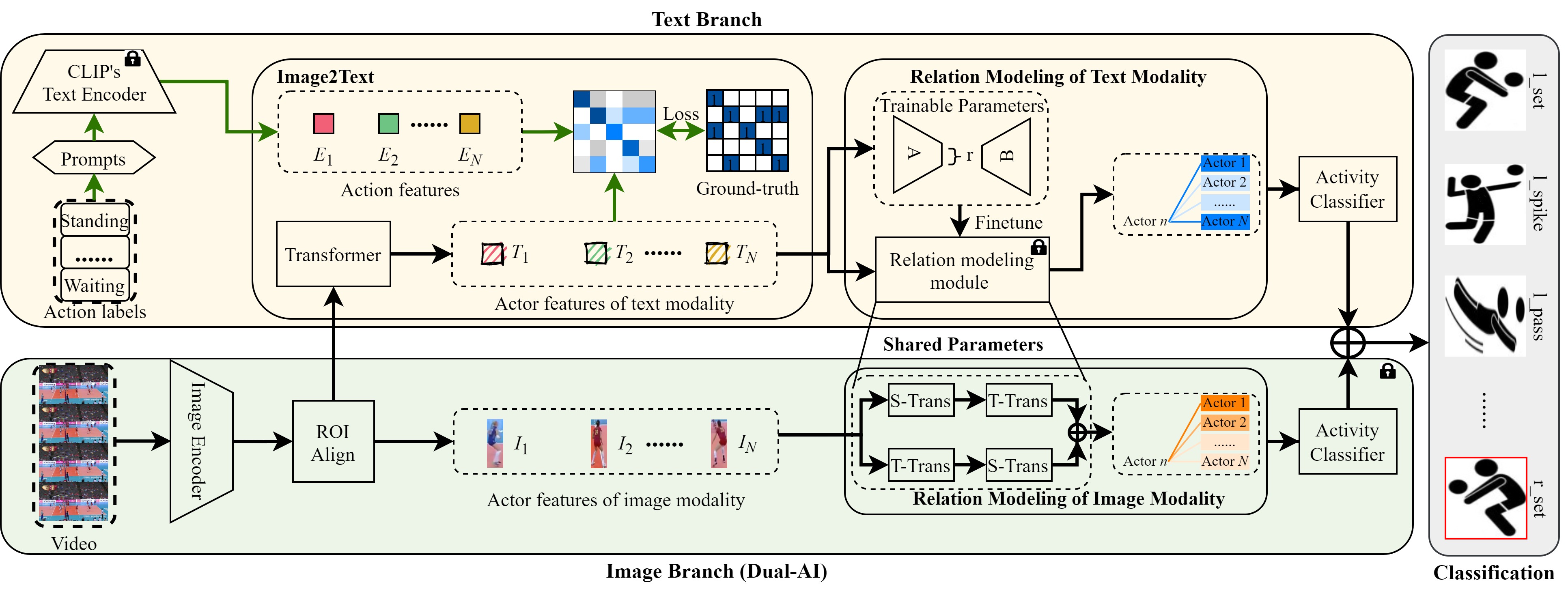}\\
\caption{Overview of the ActivityCLIP. \textbf{The process indicated by the green arrow only occurs during the training stage.} Here, we use Dual-AI as the image branch to show the process of ActivityCLIP. Dual-AI employs two paths with different spatial-temporal orders (spatial-temporal and temporal-spatial paths) for interaction relation modeling. S-Trans and T-Trans represent the spatial-Transformer and temporal-Transformer, respectively. More details on Dual-AI can be found in \cite{han2022dual}.}\label{Figure 2}
\end{figure*}

\subsection{Our Approach: ActivityCLIP}
\textbf{Overview.} The ActivityCLIP aims to mine the text information contained in action labels to supplement the image information, improving the model's performance, as shown in Fig. \ref{Figure 2}. Specifically, the ActivityCLIP consists of an image branch and a text branch. The image branch is the existing image-based method, such as ARG \cite{wu2019learning}, AFormer \cite{gavrilyuk2020actor}, Dual-AI \cite{han2022dual}, etc. We propose a plug-and-play way to plug the text branch into the image branch. In the text branch, we use the prompt (More details in section 4.2.) to describe the action label and employ the CLIP's text encoder to extract the text information. Further, we use knowledge distillation as a bridge to make our proposed Image2Text module learn how to transfer image information to text information. It is worth noting that the CLIP's text encoder is only used in the training stage. In addition, we add fewer trainable parameters based on the image branch's interaction relation modeling module and finetune them to conveniently achieve interaction relation modeling of the text branch. Finally, we adopt the later fusion way that adds the classification score of both branches to recognize group activity.

In the ActivityCLIP, the backward propagation updates only the parameters of Image2Text, matrix A and B, and Activity Classifier in the text branch.

\begin{figure}[htbp]
\centering
\includegraphics[width=0.7\textwidth]{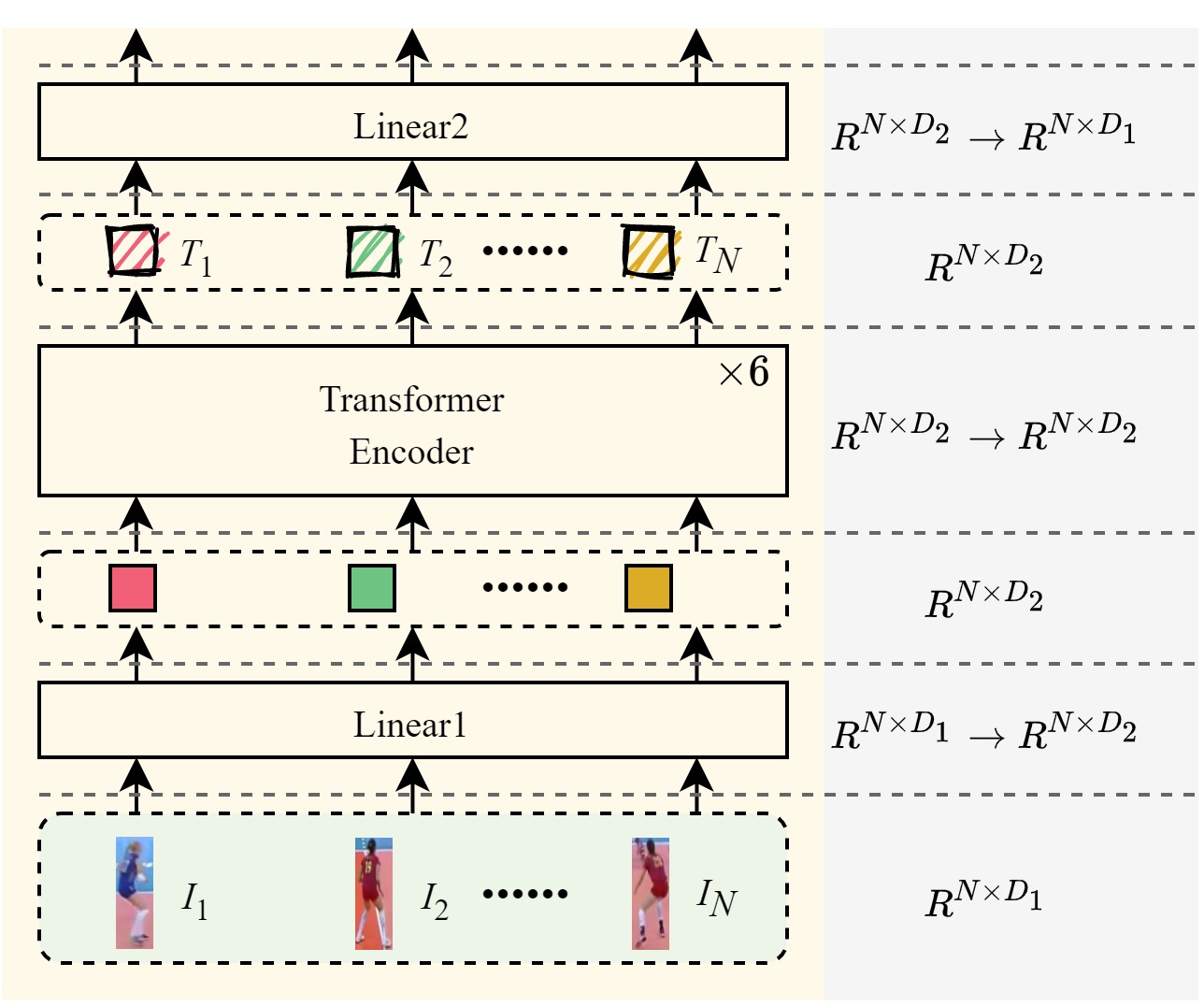}\\
\caption{The structure of Image2Text. We indicate the feature's dimensional changes in each component.}\label{Figure 3}
\end{figure}

\textbf{Image2Text.} Image2Text is crucial for ActivityCLIP. We use the `Image Encoder + Image2Text' to mine the text information. Specifically, as shown in Fig. \ref{Figure 3}, The Image2Text consists of the linear layer and transformer encoder. We use the actor's image feature $ I \in {R^{N \times D_1}} $ as input. Due to the dimension of the image feature possibly being different from the dimension of the text feature extracted by CLIP, we use Linear1 to achieve the feature’s dimensional transformation. It only includes one linear layer.

Considering the action’s co-occurrence probability of different actors in GAR tasks is important. Specifically, the co-occurrence probability of each action is different. For example, the actions of `spiking' and `blocking' often appear together in volleyball competitions. Therefore, we use the transformer encoder as the core component of Image2Text. Through the self-attention mechanism, Image2Text can pay attention to the relation of different actions, which has two advantages. First, this way can implicitly find the co-occurrence probability of different actions. Second, this way can pull the actor's features with the same semantics nearer and push the actor's features with different semantics farther. The transformer encoder's output is $ T \in {R^{N \times D_2}} $. The transformer encoder consists of 6 layers. On the one hand, the deeper transformer encoder can extract high-level features. On the other hand, the deeper structure can better absorb the distillation knowledge of CLIP.

Since we want to plug the text branch into the image branch, we must keep the dimensions of the image and text features consistent. Therefore, Linear2 aims to achieve the inverse dimensional transformation. Like Linear1, Linear2 also only includes one linear layer.

In addition, knowledge distillation is an effective approach to achieving information transfer and has been applied in many fields \cite{gou2021knowledge,li2023object,miles2023mobilevos}. Here, we use knowledge distillation to achieve information transfer between the image and text information. Like Eq. (\ref{eq1}), we use $ T $ (The output features of Image2Text) and $ E $ (The features extracted by CLIP's encoder) to compute the probability $ P_i^{E,T}(E) $ and $ P_i^{T,E}(T) $. However, one feature of $ E $ possibly corresponds to several features of $ T $. Therefore, it is not appropriate to regard the process of knowledge distillation as a 1-in-N classification problem with the CE loss. We define the KD (Knowledge Distillation) loss with the KL (Kullback-Leibler) divergence to supervise the process of information transfer, as shown in Eq. (\ref{eq3}). 

\begin{equation} \label{eq3}
\begin{array}{l}
{L_{KD}} = \frac{1}{2}[KL({P^{E,T}},{G^{E,T}}) + KL({P^{T,E}},{G^{T,E}})]
\end{array}
\end{equation}

$ {G^{E,T}} $ and $ {G^{T,E}} $ represent the ground truths, respectively. Through the information transfer, Image2Text can convert the image information into corresponding text information.

\textbf{Interaction Relation Modeling for Text Branch.} Inspired by human perception, just tell humans the action's semantics and location of each actor, and humans also can think about the actor interaction relation for recognizing group activity. Therefore, it is important to model the interaction relation of actors described by text information. On the one hand, to ensure the convenience and universality of our method, we do not want to design a new module for modeling the actor interaction relation of the text branch. On the other hand, due to the cross-modality semantic consistency of group activity, we consider that the actor interaction relations described by image and text modalities are remarkably similar. To this end, we have not trained the parameters of the interaction relation modeling module of the text branch from scratch. Based on the inspiration brought by LoRa \cite{hu2021lora}, we add fewer parameters to the interaction relation modeling module of the image branch and finetune these parameters to model the actor interaction relation of the text branch. This way has two obvious advantages. First, we only train fewer parameters, which reduces training consumption. Second, this way makes our method conveniently plugs the text branch into the image branch, improving our method’s generality.

As shown in Fig. \ref{Figure 2}, we add the matrix A and B to the pretrained parameters of the image branch's interaction relation modeling module. In the GAR task, the existing methods usually use the Transformer or GCN (Graph Convolutional Network) for actor interaction relation modeling. In transformer-based methods, we add the matrix A and B to the Q (Query), K (Key), and V (Value), respectively. In GCN-based methods, such as ARG, we add both matrices to all linear layers of the interaction relation modeling module. 

As shown in Eq.(\ref{eq4}), we update features by the low-rank matrix A and B. Where, $ {W_0} \in {R^{M \times M}} $, $ B \in {R^{M \times r}} $, and $ A \in {R^{r \times M}} $. $ W_0 $ is the pre-trained parameters and froze in the training stage. $ \alpha $ is the ratio to control the influence of both low-rank matrices.

\begin{equation} \label{eq4}
\begin{array}{l}
F(x) = {W_0}(x) + \alpha BA(x)
\end{array}
\end{equation}

\section{Experiments}

In this section, we conduct a series of experiments to evaluate ActivityCLIP’s effectiveness. First, we introduce detailed information on the datasets and evaluation metrics. Second, we introduce the implementation details of ActivityCLIP. Third, we compare our method with state-of-the-art methods to verify ActivityCLIP’s performance. Fourth, we conduct ablation studies to analyze the influence of each part. Finally, we visualize the actor interaction relation of image and text branches to analyze their differences.

\subsection{Dataset and Evaluation Metric}

\textbf{Dataset.} Following \cite{wu2019learning,gavrilyuk2020actor,han2022dual,xie2023actor,lang2023knowledge}, we use the Volleyball and Collective Activity datasets as benchmarks. \textbf{(1) The Volleyball dataset} \cite{ibrahim2016hierarchical} belongs to the volleyball competition scene. It consists of 4830 clips, of which 3493 are used for training, and the remaining are used for testing. Each label (actor’s bounding box, individual action labels, and group activity labels) was annotated on the middle frame of each clip. The individual action labels include waiting, setting, digging, falling, spiking, blocking, jumping, moving, and standing. Further, the group activity labels consist of right set, right spike, right pass, right win-point, left set, left spike, left pass, and left win-point. \textbf{(2) The Collective Activity dataset} \cite{choi2009they} belongs to the daily life scene. It includes 44 short video sequences. Like \cite{han2022dual,li2022learningAction}, we use two-thirds of the videos for training and the remaining for testing. This dataset was also annotated on the center frame of every 10th frame, using the actor’s bounding box, individual action labels, and group activity labels. The individual action labels include NA, crossing, waiting, queuing, walking, and talking. The activity labels include crossing, waiting, queuing, walking, and talking. The activity label is determined by each clip’s maximum number of action labels. Like \cite{yan2018participation,9241410HiGCIN,yuan2021spatio,han2022dual}, we merge the crossing and walking into moving.

\textbf{Evaluation Metric.} To fully evaluate ActivityCLIP's performance, we adopt the MCA (Multi-Class Accuracy), MPCA (Mean Per-Class Accuracy), and confusion matrix as evaluation metrics. Many methods have widely used these metrics \cite{wu2019learning,gavrilyuk2020actor,han2022dual,yuan2021spatio}. The MCA represents the percentage of correctly classified instances in all categories. The MPCA indicates the mean accuracy of all categories. The confusion matrix aims to analyze the model’s performance in each category deeply.

\subsection{Implementation Details}

Since ActivityCLIP is a plug-and-play approach, we replicated three representative methods (ARG, AFormer, and Dual-AI) to evaluate ActivityCLIP’s generality. We reproduced the ARG using their public code \cite{wu2019learning}. We replicated the AFormer using the public code in \cite{gavrilyuk2020actor}. We replicated the Dual-AI by referring to the detailed introduction in \cite{han2022dual}. We named the reproduced methods by using ActivityCLIP as ACLIP(ARG), ACLIP(AFormer), and ACLIP(Dual-AI), respectively. Additionally, like \cite{wang2023actionclip}, we defined $ K=11 $ discrete manual prompts, which can be classified into three types: prefix prompt (such as, `\underline{action label} this is an action'), suffix prompt (such as, `Playing action of \underline{action label}') and action label-only (such as, `\underline{action label}'). 

In the Volleyball dataset, similar to \cite{wu2019learning,han2022dual,yuan2021spatio}, we randomly sampled three frames for training and nine frames for testing in ACLIP(ARG) and ACLIP(Dual-AI), and we used ten frames for both training and testing in ACLIP(AFormer). In all three methods, we train our model in 30 epochs using a batch size of 8 and a learning rate of 0.0001. In addition, we adopt the AdamW to optimize our model with the warmup mechanism, where the epoch of warmup is 5. For the ACLIP(ARG), ACLIP(AFormer), and ACLIP(Dual-AI), we set $ \alpha $ to 8 and set $ r $ to 7, 6, and 2, respectively. The other details in the Collective Activity dataset are the same as those in the Volleyball dataset, except for batch size, which we set to 16. Finally, we conduct all experiments using the PyTorch deep learning framework in 2 Nvidia GeForce RTX 3090 GPUs.

\begin{table}[htbp]
  \footnotesize
  \setlength\tabcolsep{2.5pt}
  \centering
  \caption{Compared ActivityCLIP with baselines. We conducted experiments on the Volleyball dataset and counted the adding trainable parameters on the base of the image branch. `*' represents the baseline in which the image and text branches are combined in series. Specifically, we first use the image branch to recognize the result of action and group activity. Then, we select the corresponding action labels based on the image branch’s action result as the text branch’s input to recognize group activity again, where the text branch uses CLIP's encoder to extract the text information. Finally, both branch’s activity scores are fused to obtain the final group activity results.}
  \begin{tabular}{ccll}
\hline
Method          & \begin{tabular}[c]{@{}c@{}}Trainable\\ Parameters\end{tabular} & \begin{tabular}[c]{@{}c@{}}Group Activity\\ (MCA)\end{tabular} \\ \hline
ARG             & -                                                                                                                        & \quad 92.3                                                          \\
ACLIP*(ARG)     & 26.90M                                                                                                                   & \quad \textbf{93.0} \textcolor{red}{$\uparrow$ 0.7}                                                          \\
ACLIP(ARG)      & 9.49M                                                                                                                    & \quad \textbf{93.0} \textcolor{red}{$\uparrow$ 0.7}                                                          \\ \hline
AFormer         & -                                                                                                                        & \quad 92.2                                                          \\
ACLIP*(AFormer) & 0.46M                                                                                                                    & \quad 92.1 \textcolor[RGB]{0,190,0}{$\downarrow$ 0.1}                                                         \\
ACLIP(AFormer)  & 8.17M                                                                                                                    & \quad \textbf{92.5} \textcolor{red}{$\uparrow$ 0.3}                                                          \\ \hline
Dual-AI         & -                                                                                                                        & \quad 93.8                                                          \\
ACLIP*(Dual-AI) & 1.46M                                                                                                                    & \quad 93.2 \textcolor[RGB]{0,190,0}{$\downarrow$ 0.6}                                                          \\
ACLIP(Dual-AI)  & 8.18M                                                                                                                    & \quad \textbf{94.2} \textcolor{red}{$\uparrow$ 0.4}                                                          \\ \hline
\end{tabular}
   \label{Table 1} 
 \end{table}

\subsection{Comparison with the State-of-the-Art Methods} 

\textbf{Quantitative Analysis with Baselines:} To verify the generality of our method, we reproduced three representative methods by ActivityCLIP, as shown in Table \ref{Table 1}. Additionally, we construct another baseline, ACLIP*, in which image and text branches are combined in series.

In ARG-based methods, ACLIP*(ARG) and ACLIP(ARG) can get the equivalent performance (93.0 vs. 93.0), outperforming ARG with a noticeable gap. In AFormer-based methods, ACLIP(AFprmer) has an increase of 0.3 (92.5 vs. 92.2) compared to AFormer. However, ACLIP*(AFormer) decreases by 0.1 (92.1 vs. 92.2). In Dual-AI-based methods, ACLIP(Dual-AI) also outperforms Dual-AI by 0.4 (94.2 vs. 93.8). Compared with Dual-AI, ACLIP*(AFormer) shows an obvious performance degradation with 0.6 (93.2 vs. 93.8) again. The above analyses show that ActivityCLIP demonstrates favorable generality in various methods. Nevertheless, the performance of ACLIP* is unsatisfactory.

We have summarized the following two reasons to explain this phenomenon. First, through Knowledge distillation, the Image2Text module in ACLIP has the capacity to mine the text information, enabling our method to adapt naturally to various methods. Second, the ACLIP* needs to rely on the action result of the image branch to select the corresponding action labels to mine the text information by the CLIP. Therefore, if the action results of the image branch are not accurate enough, the text branch of the ACLIP* will mine the invalid information.

Additionally, the main trainable parameters of ACLIP and ACLIP* are from the Image2Text and interaction relation modeling modules, respectively. Thus, due to the large number of parameters in ARG's interaction relation module, the number of trainable parameters of ACLIP*(ARG) is higher than ACLIP(ARG). Then, compared with ACLIP* (AFormer) and ACLIP*(Dual-AI), ACLIP(AFormer) and ACLIP(Dual-AI) have a higher number of trainable parameters. We think adding appropriate trainable parameters to achieve more robust performance is worthwhile.

\begin{table*}[ht]
  \footnotesize
  \setlength\tabcolsep{2pt}
  \centering
  \caption{Compared ACLIP(Dual-AI) with state-of-the-art methods on the Volleyball and Collective Activity datasets. `$\_$' indicates that this result cannot be obtained from their paper. The best performance is shown in \textbf{bold} and the second best is shown in \underline{underline}.}
  \begin{tabular}{ccccccc}
\hline
Method            & Backbone     & \multicolumn{2}{c}{Volleyball Dataset} &  & \multicolumn{2}{c}{Collective Activity Dataset} \\ \cline{3-4} \cline{6-7} 
                  &              & MCA                & MPCA              &  & MCA                    & MPCA                   \\ \hline
StagNet(2020)\cite{qi2019stagnet}     & VGG-16       & 89.3               & -                 &  & -                      & -                      \\
HiGCN(2020)\cite{9241410HiGCIN}       & ResNet-18    & 91.4               & 92.0              &  & 93.4                   & 93.0                   \\
PRL(2020)\cite{hu2020progressive}         & VGG-16       & 91.4               & 91.8              &  & -                      & 93.8                   \\
DIN(2021)\cite{yuan2021spatio}         & VGG-16       & 93.6               & 93.8              &  & -                      & 95.9                   \\
P$^2$CTDM(2021)\cite{yan2021position}      & Inception-v3 & 91.8               & 92.7              &  & -                      & 96.1                   \\
TCE+STBiP(2021)\cite{yuan2021learning}   & Inception-v3 & 93.3               & 93.4              &  & -                      & 96.4                   \\
GroupFormer(2021)\cite{li2021groupformer} & Inception-v3 & 94.1               & -                 &  & -                      & -                      \\
ASTFormer(2022)\cite{li2022learningAction}   & Inception-v3 & 93.5               & 93.9              &  & 96.5                   & 95.3                   \\
CCG-LSTM(2022)\cite{tang2022coherence}    & AlexNet      & -                  & -                 &  & -                      & 93.0                   \\
COMPOSER\cite{zhou2022composer}          & HRNet        & -                  & -                 &  & 96.2                   & -                      \\
3DUSTG(2023)\cite{wang20233d}      & I3D          & 93.2               & -                 &  & -                      & -                      \\
HSTT(2023)\cite{zhu2023hierarchical}        & Inception-v3 & 93.7               & -                 &  & 96.1                   & -                      \\
KRGFormer(2023)\cite{pei2023key}   & Inception-v3 & \textbf{94.6}               & \textbf{94.8}              &  &                        &                        \\
LG-CAGR(2024)\cite{JIANG2024108412}     & VGG-19       & 92.1               & 92.4              &  & -                      & 97.1                   \\
ACLIP(Dual-AI)    & Inception-v3 & \underline{94.2}               & \underline{94.4}              &  & \textbf{97.9}                   & \textbf{97.3}                   \\ \hline
\end{tabular}
   \label{Table 2}
 \end{table*}

\textbf{Quantitative Analysis with other methods on the Volleyball Dataset:} To verify the performance of our method, we conduct comprehensive comparisons with existing methods, including various backbones, such as ResNet-18 \cite{he2016deep}, VGG-16/19 \cite{simonyan2014very}, I3D \cite{carreira2017quo}, and Inception-v3 \cite{szegedy2016rethinking}. As shown in Table \ref{Table 2}, except for KRGFormer, ACLIP(Dual-AI) performs better than other methods. Compared with 3DUSTG, ACLIP(Dual-AI) has an increase of 1.0 (94.2 vs. 93.2) under the metric MCA. This indicates that interaction relation modeling is crucial for this task, even using a stronger backbone for feature extraction. Then, compared with HSTT, ACLIP(Dual-AI) increases by 0.5 (94.2 vs. 93.7) under the metric MCA. Further, ACLIP(Dual-AI) also has a performance improvement of 2.1 (94.2 vs. 92.1) and 2.0 (94.4 vs. 92.4) under the metrics MCA and MPCA than LG-CAGR. These prove that ACLIP(Dual-AI) extracts more useful interaction relations for recognizing group activity. However, compared with KRGFormer, ACLIP(Dual-AI) has a decrease of 0.4 (94.2 vs. 94.6) and 0.4 (94.4 vs. 94.8) under the metrics MCA and MPCA, respectively. This is because they used data augmentation to augment individual patches, which increases the model’s performance with 0.7 and 0.6 under the metrics MCA and MPCA, respectively. Our method only used the original data for training.

The above analysis shows that ActivityCLIP is a promising method that can mine complementary information to help existing methods achieve more competitive performance.

\textbf{Quantitative Analysis with other methods on the Collective Activity Dataset:} As shown in Table \ref{Table 2}, ACLIP(Dual-AI) obtains a favorable performance compared to other methods. Compared with these methods (PRL, DIN, P$^2$CTDM, and TCE+STBiP), ACLIP (Dual-AI) improves the model's performance with a noticeable gap. Compared with COMPOSER, ACLIP(Dual-AI) increases by 1.7 (97.9 vs. 96.2) under the metric MCA, even if they use HRNet to mine the skeleton information. Then, compared with LG-CAGR, ACLIP(Dual-AI) also has a performance improvement of 0.2 (97.3 vs. 97.1) under the metric MPCA.

Based on these analyses, we observe that our method can obtain competitive performance in various scenarios. This is attributed to the improved robustness of the model with the support of text information.

 \begin{figure*}[ht]
  \centering
  \includegraphics[width=1.0\textwidth]{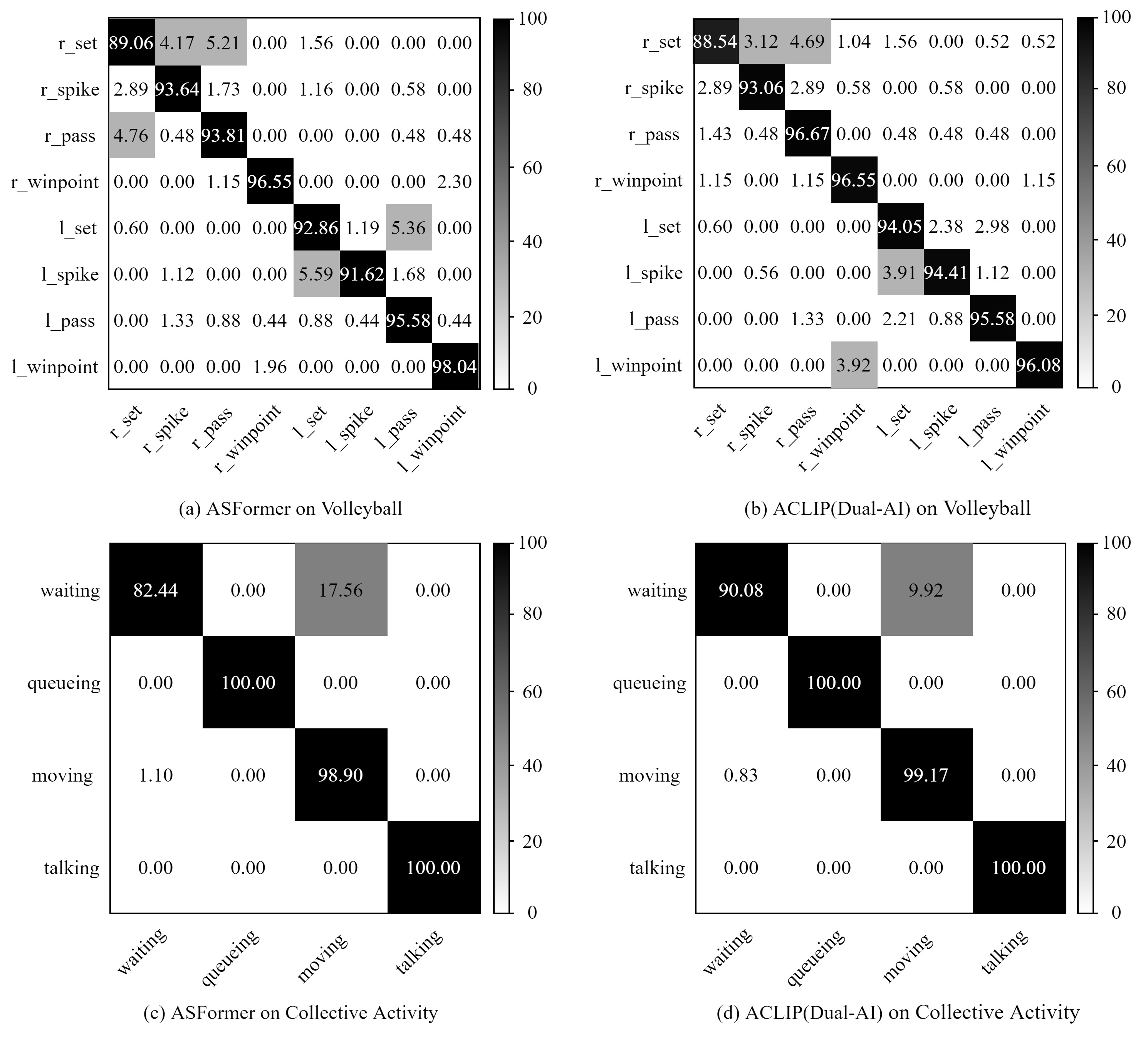}\\
  \caption{The confusion matrix analysis on the ASTFormer \cite{li2022learningAction} and ACLIP(Dual-AI). Figures (a) and (b) represent the confusion matrix of ASTFormer and ACLIP(Dual-AI), respectively, on the Volleyball dataset; Figures (c) and (d) represent the confusion matrix of both methods, respectively, on the Collective Activity dataset. Here, `r' and `l' are short for `right' and `left.'}\label{Figure 4}
 \end{figure*}

\textbf{Confusion Matrix Analysis:} To analyze the model’s performance deeply, we use the confusion matrix to show the classification situation of ACLIP(Dual-AI) and ASTFormer in each category. 

As shown in Fig. \ref{Figure 4} (a) and (b), ACLIP(Dual-AI) outperforms the ASFormer with obvious improvement in three categories (r$\_$pass, l$\_$set, and l$\_$spike). Specifically, ACLIP(Dual-AI) reduce the confusion of `r$\_$pass’ and `r$\_$set’, `l$\_$set’ and `l$\_$pass’, `l$\_$spike’ and `l$\_$set.’ We think that ACLIP(Dual-AI) can provide rich information to describe the actor's actions, which is beneficial for modeling more accurate actor interaction relations to distinguish different group activities. However, ACLIP(Dual-AI) increases the confusion between `l$\_$winpoint’ and `r$\_$winpoint.’ This is caused by the failure to distinguish the group activity's location correctly. Therefore, in the future, we need to explore more suitable solutions to embed the actor’s position information into the text branch.

As shown in Fig. \ref{Figure 4}(c) and (d), like ASFormer, all instances in two categories (queueing and talking) are classified accurately by ACLIP(Dual-AI). Meanwhile, ACLIP(Dual-AI) also outperforms ASFormer in other categories (waiting and moving).

\subsection{Ablation Studies}

\begin{table}[htbp]
  \footnotesize
  \setlength\tabcolsep{2.5pt}
  \centering
  \caption{Ablation study of ActivityCLIP's Image2Text. We conducted these experiments on the Volleyball dataset and counted the parameters of Image2Text.}
  \begin{tabular}{cccc}
\hline
Layer Number & $ {L_{KD}} $ & Parameters  & \begin{tabular}[c]{@{}c@{}}Group Activity\\ (MCA)\end{tabular} \\ \hline
0            & $\surd$   & 0.26M  & 93.9                                                          \\
2            & $\surd$   & 2.89M  & 93.9                                                          \\
4            & $\surd$   & 5.52M  & 93.6                                                          \\
6            & $\surd$   & 8.16M  & \textbf{94.2}                                                          \\
8            & $\surd$   & 10.79M & 93.7                                                          \\
10           & $\surd$   & 13.42M & 93.8                                                          \\ \hline
6            &     & 8.16M  & 93.6                                                          \\ \hline
\end{tabular}
   \label{Table 3} 
 \end{table}

\textbf{1) Ablation Studies for the Image2Text.} We conduct several experiments to verify the effectiveness of Image2Text from two aspects, as shown in Table \ref{Table 3}. First, we verify the performance improvement's root cause, conducting based on  $ Layer Number = 6 $. With the supervision of $ {L_{KD}} $, we observe an increase of 0.6 (94.2 vs. 93.6). This indicates that the text modality provides complementary information to the image modality for recognizing group activity instead of the performance improvement brought by the additional parameters.

Second, we attempt to investigate the impact of the transformer encoder’s layer number in Image2Text. The transformer encoders are stacked in a concatenated manner with different layer numbers. We observe that when $ Layer Number = 6 $, our method gets the best performance. On the one hand, Image2Text needs a suitable amount of parameters to store the text information transferred by CLIP. Thus, little layer numbers cannot achieve good results. On the other hand, more layer numbers make it more challenging to optimize our model. In the subsequent experiments, we use the setting of $ Layer Number = 6 $.

\begin{table}[htbp]
  \footnotesize
  \setlength\tabcolsep{2.5pt}
  \centering
  \caption{Ablation study of text modality's interaction relation modeling module. We conducted these experiments on the Volleyball dataset and counted the trainable parameters of this module. $\dag$ indicates that we don't plug the text branch's interaction relation modeling module into the image branch. Specifically, in the text branch, we use the same interaction relation modeling module of the image branch. Then, we train the text branch's interaction relation module from scratch.}
  \begin{tabular}{ccc}
\hline
Method          & \begin{tabular}[c]{@{}c@{}}Trainable\\ Parameters\end{tabular}  & \begin{tabular}[c]{@{}c@{}}Group Activity\\ (MCA)\end{tabular} \\ \hline
ACLIP(ARG)\dag     & 26.35M & \textbf{93.0}                                                          \\
ACLIP(ARG)      & 0.52M  & \textbf{93.0}                                                          \\ \hline
ACLIP(AFormer)\dag & 0.33M  & \textbf{92.5}                                                          \\
ACLIP(AFormer)  & 0.01M  & \textbf{92.5}                                                          \\ \hline
ACLIP(Dual-AI)\dag & 1.30M   & \textbf{94.2}                                                          \\
ACLIP(Dual-AI)  & 0.01M  & \textbf{94.2}                                                          \\ \hline
\end{tabular}
   \label{Table 4} 
 \end{table}

\textbf{2) Ablation Studies for the Effectiveness of the Text Branch's Interaction Relation Modeling Module.} We propose to plug the text branch’s interaction relation modeling into the image branch. To verify the effectiveness of this method, we conducted experiments using three representative methods, as shown in Table \ref{Table 4}. On the ARG-based methods, we observe that ACLIP(ARG) only uses 1.97$\%$ trainable parameters of ACLIP(ARG)$\dag$ to get an identical performance (93.0 vs. 93.0). Then, on the AFormer-based method, the performance of ACLIP(AFormer) and ACLIP(AFormer)$\dag$ is consistent (92.5 vs. 92.5), but ACLIP (AFormer) only uses 3.03$\%$ trainable parameters of ACLIP (AFormer)$\dag$. Further, on the Dual-AI-based method, we note that ACLIP-(Dual-AI) uses 0.77$\%$ trainable parameters of ACLIP(Dual-AI)$\dag$ to obtain the same performance (94.2 vs. 94.2). 

Through the analysis of the above experiments, we have gained the following conclusions: The interaction relations of the text branch are similar to those of the image branch. Therefore, it is unnecessary to train the text branch’s interaction relation modeling module from scratch. We use fewer trainable parameters based on the image branch’s interaction relation modeling module to achieve the interaction relation modeling of the text branch, which can still achieve promising performance.

\textbf{3) Ablation Studies for the Hyper-parameters of the Text Branch's Interaction Relation Modeling Module.} Inspired by LoRa, we plug the text branch's interaction relation modeling module into the image branch. This process is controlled by \textit{r} and $ \alpha $. The former can control the number of trainable parameters; the latter is the ratio to control the influence of trainable parameters. As shown in Table \ref{Table 5}, simply increasing \textit{r} (i.e., increasing the number of trainable parameters) does not improve performance. Additionally, when $ \alpha = 8 $, our method can gain the best performance.

These observations are reasonable. Since the interaction relations of both branches are similar, adding more parameters may damage the capacity of the interaction relation modeling learned in the image branch. Thus, we only add fewer trainable parameters to learn text branch-specific interaction relations. Similarly, we also chose a reasonable $\alpha$ to control the influence of trainable parameters.

In other methods, such as ACLIP(ARG) and ACLIP-(AFormer), due to their use of different interaction relation modeling modules, we adjust the $r$ (See more details in section 4.2 on hyper-parameters setting.) on different methods. Further, we use the same $\alpha$ in all methods.

\begin{table}[htbp]
  \footnotesize
  \setlength\tabcolsep{3.5pt}
  \centering
  \caption{Ablation study of the hyper-parameters on the text branch's interaction relation modeling module. We conducted these experiments using ACLIP(Dual-AI) on the Volleyball dataset and counted the trainable parameters in this module.}
  \begin{tabular}{cccc}
\hline
                         & $ r $     & \begin{tabular}[c]{@{}c@{}}Trainable\\ Parameters\end{tabular} & \begin{tabular}[c]{@{}c@{}}Group Activity\\ (MCA)\end{tabular} \\ \hline
\multirow{6}{*}{$ \alpha=8 $} & 1     & 6.14K                                                          & 93.9                                                          \\
                         & 2     & 12.29K                                                         & \textbf{94.2}                                                          \\
                         & 4     & 24.58K                                                         & 93.9                                                          \\
                         & 6     & 36.86K                                                         & 94.0                                                          \\
                         & 8     & 49.15K                                                         & 93.9                                                          \\
                         & 10    & 61.44K                                                         & 93.8                                                          \\ \hline
                         & $ \alpha $ & \begin{tabular}[c]{@{}c@{}}Trainable\\ Parameters\end{tabular} & \begin{tabular}[c]{@{}c@{}}Group Activity\\ (MCA)\end{tabular} \\ \hline
\multirow{6}{*}{\textit{r}=2}     & 1     & 12.29K                                                         & 94.0                                                          \\
                         & 2     & 12.29K                                                         & 94.1                                                          \\
                         & 4     & 12.29K                                                         & 93.6                                                          \\
                         & 6     & 12.29K                                                         & 94.0                                                          \\
                         & 8     & 12.29K                                                         & \textbf{94.2}                                                          \\
                         & 10    & 12.29K                                                         & 93.9                                                          \\ \hline
\end{tabular}
   \label{Table 5} 
 \end{table}

 \begin{figure}[ht]
  \centering
  \includegraphics[width=0.5\textwidth]{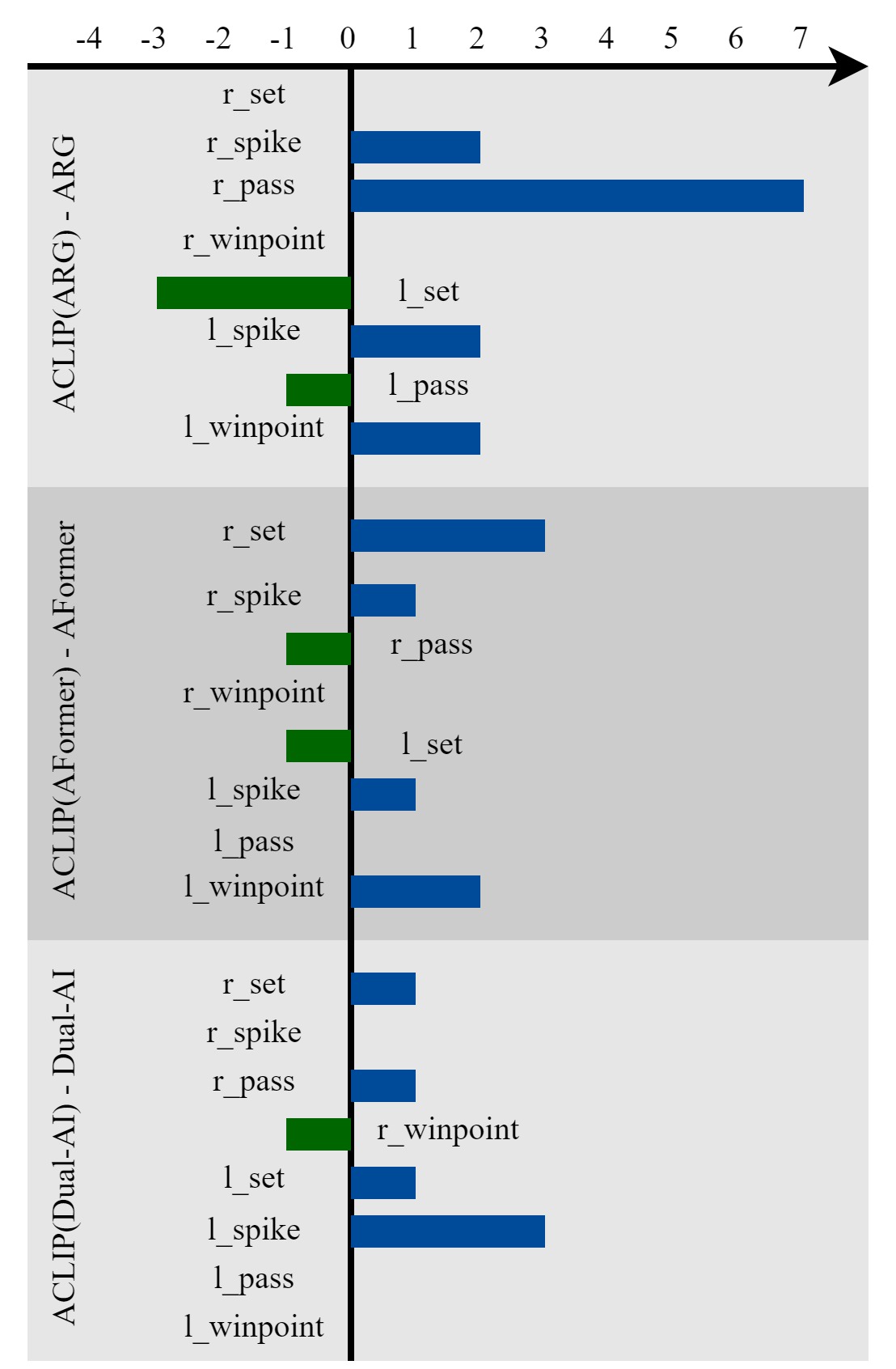}\\
  \caption{The influence of text information on each activity category. We conducted these experiments on the Volleyball dataset. `ACLIP(ARG) - ARG' represents using the number of correctly identified activities by ACLIP(ARG) in each category to subtract the number of correctly identified activities by ARG in the corresponding category. `ACLIP(AFormer) - AFormer' and `ACLIP(Dual-AI) - Dual-AI' represent the same meaning as `ACLIP(ARG) - ARG'. Here, `r' and `l' are short for `right' and `left.'}\label{Figure 5}
 \end{figure}

\textbf{4) Ablation Studies for the Influence of Text Information on Each Activity Category.} To investigate whether text information only improves the recognition performance for specific group activities, we analyze the results of our method in each activity category, as shown in Fig. \ref{Figure 5}. On the one hand, each method (ACLIP(ARG), ACLIP(AFormer), and ACLIP(Dual-AI)) all mine the effective text information to supplement the image information for achieving better group activity recognition on half of all activity categories. On the other hand, in the three methods, the text information has improved the recognition performance for a total of 6 activity categories (r$\_$set, r$\_$spike, r$\_$pass, l$\_$set, l$\_$spike, and l$\_$winpoint). These analyses indicate that text information is helpful in different activity categories.

\subsection{Visualization}

 \begin{figure*}[htbp]
  \centering
  \includegraphics[width=1.0\textwidth]{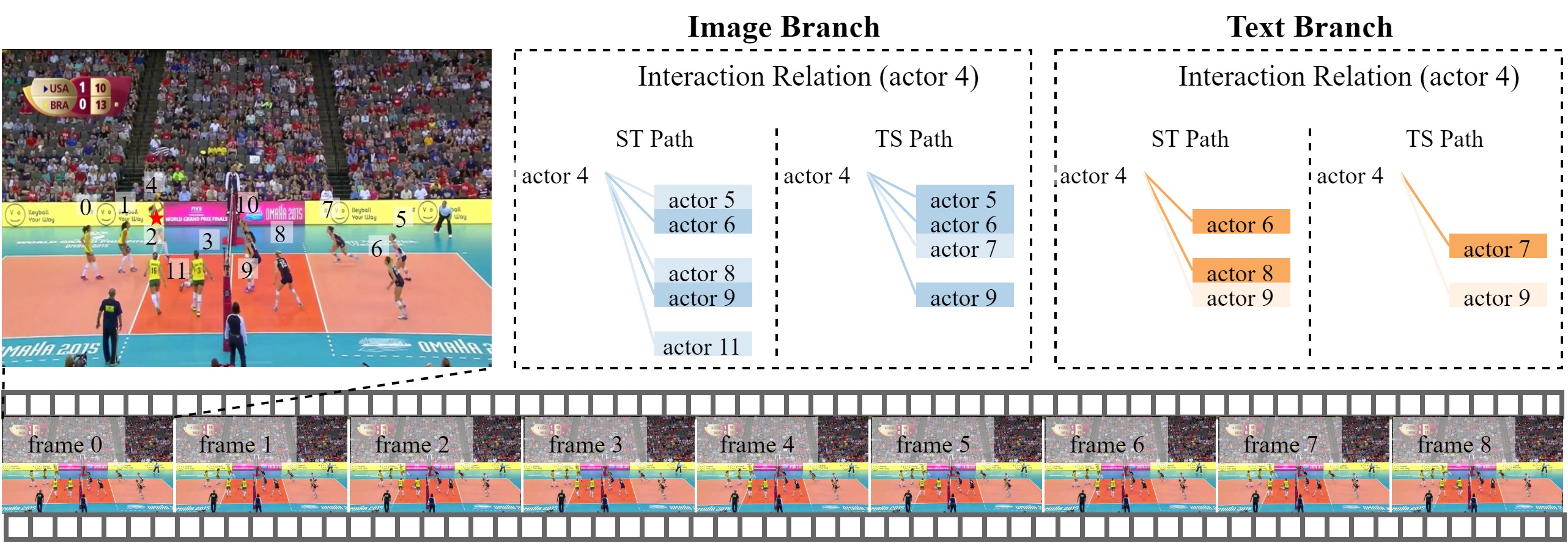}\\
  \caption{Visualizing actor’s interaction relation on both branches of ACLIP(Dual-AI). The group activity of this instance is `l$\_$spike,’ and actor 4 is the critical actor in executing this group activity. In addition, the Dual-AI’s relation modeling module consists of ST (Spatial-Temporal) and TS (Temporal-Spatial) paths. Thus, we visualize the interaction relation in all paths of actor 4 with other actors. Actor 4 is marked with a pentagram. A darker color indicates a more vital interaction relation between two actors.}\label{Figure 6}
 \end{figure*}

To verify the difference in the interaction relation of image and text branches, we visualize the interaction relation between actor 4 and other actors, as shown in Fig. \ref{Figure 6}. We find three interesting phenomena. First, the interested actors by actor 4 in the ST and TS paths of the text branch also appear in the corresponding path of the image branch. Second, actor 4 of the text branch mines the different interaction relations with the image branch and tends to focus on fewer interaction relations. Third, the group activity label of this instance is ’l$\_$spike.’ The image branch misclassified this instance as ’l$\_$set,’ while the text branch outputs the correct classification result. With the support of the text branch, ACLIP(Dual-AI) ultimately obtains the correct classification result.

We have carefully analyzed the reasons for the observed phenomena. First, we plug fewer trainable parameters into the image branch to model the text branch's interaction relation, which makes the text branch's actor 4 focus on similar actors with the image branch. Second, with the support of text information, actor 4 models the more important interaction relations and ignores irrelevant interaction relations. Third, the text branch provides complementary information to correct the error of the image branch, which helps the model obtain correct classification results.

\section{Conclusion}

We propose ActivityCLIP, which mines text information to supplement the image information for enhancing group activity recognition. ActivityCLIP is a plug-and-play method. Specifically, we propose the Image2Text module to learn the ability of information transfer under the guidance of CLIP. We also plug the text branch's interaction relation modeling module into the image branch using only fewer trainable parameters. Through the above ways, ActivityCLIP can conveniently be applied to various image-based methods. We reproduce three representative methods by ActivityCLIP to prove the generality of our method. Then, we compare our method with the state-of-the-art methods to show the favorable performance of ActivityCLIP. Finally, we conducted a series of ablation studies to verify the effectiveness of each module and hyper-parameter.

\section{Acknowledgments}

This work was supported partly by the National Natural Science Foundation of China (Grant No. 62173045, 62273054), partly by the Fundamental Research Funds for the Central Universities (Grant No. 2020XD-A04-3), and the Natural Science Foundation of Hainan Province (Grant No. 622RC675).

\bibliographystyle{elsarticle-num}
\bibliography{mybibfile}

\clearpage
\clearpage

\end{document}

%% file: math_def.tex
\newcommand{\ttinneriter}{\texttt{inner\_iters}}
\newcommand{\ttit}{\texttt{iter}}
\newcommand{\ttmaxiter}{\texttt{max\_iter}}
\newcommand{\ttbestScore}{\texttt{best\_score}}
\newcommand{\ttbestR}{\texttt{best\_rotation}}
\newcommand{\ttbestT}{\texttt{best\_translation}}

\newcommand{\cL}{\mathcal{L}}
\newcommand{\cM}{\mathcal{M}}
\newcommand{\cN}{\mathcal{N}}
\newcommand{\cI}{\mathcal{I}}
\newcommand{\cS}{\mathcal{S}}
\newcommand{\cD}{\mathcal{D}}
\newcommand{\cP}{\mathcal{P}}
\newcommand{\cQ}{\mathcal{Q}}
\newcommand{\cO}{\mathcal{O}}
\newcommand{\cT}{\mathcal{T}}
\newcommand{\cad}{\mathcal{d}}
\newcommand{\cX}{\mathcal{X}}
\newcommand{\cXh}{\hat{\mathcal{X}}}
\newcommand{\cC}{\mathcal{C}}
\newcommand{\cF}{\mathcal{F}}

\newcommand{\be}{\mathbf{e}}
\newcommand{\br}{\mathbf{r}}
\newcommand{\bx}{\mathbf{x}}
\newcommand{\bxh}{\hat{\mathbf{x}}}
\newcommand{\bX}{\mathbf{X}}
\newcommand{\bY}{\mathbf{Y}}
\newcommand{\bZero}{\mathbf{0}}
\newcommand{\hbX}{\hat{\mathbf{X}}}
\newcommand{\bS}{\mathbf{S}}
\newcommand{\bs}{\mathbf{s}}
\newcommand{\bp}{\mathbf{p}}
\newcommand{\bq}{\mathbf{q}}
\newcommand{\bD}{\mathbf{D}}
\newcommand{\bd}{\mathbf{d}}
\newcommand{\bA}{\mathbf{A}}
\newcommand{\bR}{\mathbf{R}}
\newcommand{\bt}{\mathbf{t}}
\newcommand{\bH}{\mathbf{H}}
\newcommand{\bh}{\mathbf{h}}
\newcommand{\by}{\mathbf{y}}
\newcommand{\bz}{\mathbf{z}}
\newcommand{\bu}{\mathbf{u}}
\newcommand{\ba}{\mathbf{a}}
\newcommand{\bg}{\mathbf{g}}
\newcommand{\bo}{\mathbf{o}}
\newcommand{\bl}{\mathbf{l}}
\newcommand{\bOnes}{\mathbf{1}}
\newcommand{\bF}{\mathbf{F}}
\newcommand{\bK}{\mathbf{K}}
\newcommand{\bI}{\mathbf{I}}
\newcommand{\tdf}{\tilde{f}}
\newcommand{\tdh}{\tilde{h}}

\newcommand{\bbR}{\mathbb{R}}
\newcommand{\bbE}{\mathbb{E}}
\newcommand{\bbD}{\mathbb{D}}
\newcommand{\bbF}{\mathbb{F}}
\newcommand{\bbFh}{\hat{\mathbb{F}}}
\newcommand{\bmu}{\boldsymbol{\mu}}
\newcommand{\bhr}{\hat{\mathbf{r}}}
\newcommand{\bJ}{\mathbf{J}}
\newcommand{\Nsample}{$N_\text{sample}$}

\newcommand{\kernel}{\psi}

\newcommand{\residual}{\mathbf{r}}
\newcommand{\btheta}{\boldsymbol{\theta}}